\documentclass[conference,anonymous]{IEEEtran}
\IEEEoverridecommandlockouts
\usepackage{tabularx}
\usepackage{multirow}
\usepackage[utf8]{inputenc}
\usepackage{subfigure}
\usepackage{cite}
\usepackage{booktabs}
\usepackage{amsmath,amssymb,amsfonts}
\usepackage{algorithmic}
\usepackage{graphicx}
\usepackage{textcomp}
\usepackage{xcolor}
\usepackage{pifont}
\let\oldding\ding
\renewcommand{\ding}[2][1]{\scalebox{#1}{\oldding{#2}}}
\def\BibTeX{{\rm B\kern-.05em{\sc i\kern-.025em b}\kern-.08em
    T\kern-.1667em\lower.7ex\hbox{E}\kern-.125emX}}
    \newcolumntype{C}[1]{>{\centering\let\newline\\\arraybackslash\hspace{0pt}}m{#1}}
\author{No Author}

\makeatletter
  \ifCLASSOPTIONpeerreview
    
  \fi
\makeatother

\author{Shalini Pandey, Jaideep Srivastava}
\begin{document}
\title{MOOCRep: A Unified Pre-trained Embedding of MOOC Entities }
\maketitle
\begin{abstract}
Many machine learning models have been built to tackle information overload issues on Massive Open Online Courses (MOOC) platforms. These models rely on learning powerful representations of MOOC entities. However, they suffer from the problem of scarce expert label data. To overcome this problem, we propose to learn pre-trained representations of MOOC entities using abundant unlabeled data from the structure of MOOCs which can directly be applied to the downstream tasks. While existing pre-training methods have been successful in NLP areas as they learn powerful textual representation, their models do not leverage the richer information about MOOC entities. This richer information includes the graph relationship between the lectures, concepts, and courses along with the domain knowledge about the complexity of a concept.  We develop MOOCRep, a novel  method based on  Transformer language model trained with two pre-training objectives : 1) graph-based objective to capture the powerful signal of entities and relations that exist in the graph, and 2) domain-oriented objective to effectively incorporate the  complexity level of concepts.  Our experiments reveal that MOOCRep's embeddings outperform state-of-the-art representation learning methods  on two tasks important for education community, concept pre-requisite prediction and lecture recommendation.
\end{abstract}

\maketitle

\section{Introduction}
With the development of technology, online
education has been growing rapidly over the past decade. It has provided a convenient system for people to learn at their own pace and facilitate the benefits of lifelong learning~\cite{inoue2007online}. The popularity of online learning has led to the growth of various learning platforms such as edX, Coursera, Udemy along with others. On the top, the Covid-19 pandemic has further accelerated this online learning revolution. However, with the ever-increasing amount of learning resources being added to these platforms, comes the burden of minimizing information overload. For the continued success, it is crucial to design a learning platform that can assist learners in locating and organizing learning materials in a highly efficient manner. To achieve this, it is required to design a system capable of understanding the relationship between different learning materials. It will lead to the development of tools to provide self-learners a well-curated learning path  and  create optimized curricula designs for them~\cite{tang2021conceptguide}.\par

\begin{figure}[ht]
  \centerline{\includegraphics[width=0.5\textwidth]{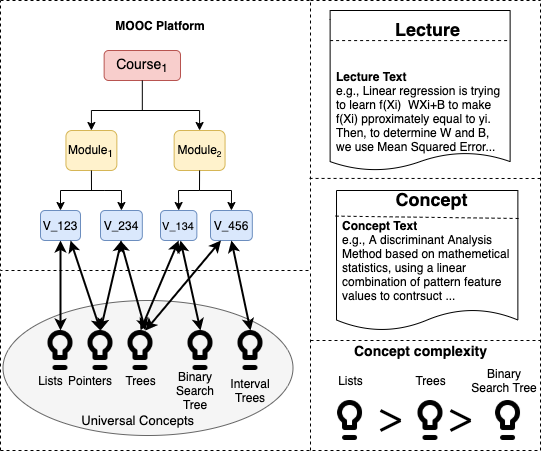}}
     \caption{An overview of data used by MOOCRep.  We use the textual content of these entities, the structure of MOOC platform and the complexity levels of concepts to learn entity representations.    }
     \label{motivation}


  \end{figure}

A MOOC platform, as shown in figure ~\ref{motivation}, provides learners with various courses from different domains. Each course covers a series of modules and each module consists of a sequence of lectures. Each lecture is also tagged with universal Wikipedia concepts which are not bound to any course; while each concept can be tagged to multiple lectures. These  lectures, courses, and concepts constitute the entities of MOOC platform and it is crucial to derive the relationships between them to improve various tasks such as, lecture recommendation~\cite{bhatt2018seqsense, zhao2018flexible}, concept path recommendation~\cite{mahapatra2018videoken}, concept pre-requisite prediction~\cite{roy2019inferring, liang2018investigating}, and course recommendation~\cite{xu2016personalized}. Multiple efforts have been put to solve these tasks. In general, the existing solutions usually design end-to-end frameworks to learn embeddings of the entities involved in the task and the objective functions are designed for minimizing task-oriented loss. However, these approaches require substantial amount of manually labeled data (e.g., concept pre-requisites), or suffer from the sparsity issue (e.g., lack of sufficiently reliable similar learners) limiting the performance of these models. \par
To tackle these issues, we propose to utilize pre-training technique~\cite{mccormick2016word2vec, beltagy2019scibert} that learns representation by leveraging the abundant unlabeled data  present in MOOC structure and textual content of entities (shown in figure~\ref{motivation}). The learned representations can then be applied directly to the downstream models. Several pre-training methods have been developed in Natural Language Processing (NLP) on representation learning ~\cite{devlin2018bert}. However, they only exploit the textual content of the entities. Simply employing these methods to learn MOOC entity representation is not sufficient. For example, consider two knowledge concepts, \textit{lists} and \textit{trees}. The textual content of the two knowledge concepts might not be semantically similar but for a student learning the \textit{data structures}, both the concepts are relevant and should be recommended together. Utilizing the rich information about the relevance relationship between the entities can benefit the model to learn effective embeddings. Additionally, some concepts on MOOC platform are advanced compared to others, for example, \textit{trees} is a basic concept required to understand advanced concept, such as, \textit{binary search tree}. Explicitly encoding the complexity  level of a concept i.e., whether the concept is a basic or an advanced one, helps the model to incorporate the ordering relations between the  concepts.\par

In this paper, we propose MOOCRep, a method to learn effective embeddings that goes beyond capturing the textual content but also incorporates the relationships between entities.
We start with  a pre-trained BERT model to learn embeddings of lectures and concepts using their textual content. Then, we use segment-aware transformer~\cite{bai2020segabert} to learn the course embedding that encodes the hierarchical structure of lectures and modules in a course shown in figure~\ref{motivation}. Further, in order to preserve the inherent relations between MOOC entities, we design pre-training objectives to train our MOOCRep model. Our first pre-training objective  captures the associated  graph relationship between  entities obtained from the MOOCCube dataset~\cite{yu2020mooccube}. We adopt a triplet loss~\cite{chechik2010large}   consisting of anchor, positive and negative samples where two entities connected by an edge in the graph  are considered as positive samples. To obtain negative samples, we employ an efficient technique introduced in~\cite{lerer2019pytorch}. The second pre-training objective incorporates domain-oriented knowledge about the complexity level of a concept. For this, we first heuristically compute the concept's complexity level based on their distributional patterns in MOOCs. Next, we adopt Mean Square Error (MSE) loss to predict the complexity levels using the concept embeddings. Training the model by combining both these losses helps in preserving the relevance between entities from the graph structure  as well as  the sequential ordering between the concepts from their complexity levels.


To evaluate our pre-trained entity embeddings, we show that the learned
representations substantially outperform the state-of-the-art models on important tasks in education domain. The two tasks we explore in this paper are concept pre-requisite prediction that can support development of concept paths for learners and next lecture recommendation to help learners find the next lecture of interest~\cite{tang2021conceptguide}. Experimental results on these two downstream tasks show the applicability and strength of our pre-trained embeddings. To summarize, our main contributions in this paper are as follows:

\begin{itemize}
    \item We are the first to introduce a method to learn generic embeddings of all MOOC entities which can be directly applied to various applications important to the education community .
    \item We develop a model that learns effective pre-trained embeddings of MOOC entities that encode their textual content along with different relationships between them.
    \item We show the improvement of pre-trained  embeddings on concept pre-requisite prediction and  lecture recommendation tasks. Our  experiments show that our pre-trained embeddings achieve on an average $8.97\%$ improvement over state-of-the-art pre-training algorithms.
\end{itemize}

\section{Related Work}
Real-world education service systems, such as massive open online courses (MOOCs) and online platforms for intelligent tutoring systems on the web offers millions of online courses which have attracted attention from the public~\cite{anderson2014engaging}. With the data collected from these systems, various data-driven techniques have been developed to improve different aspects of the platform.  Most notably, concept pre-requisite prediction~\cite{liang2018investigating, pan2017prerequisite} and lecture recommendation~\cite{pardos2017enabling,bhatt2018seqsense} have shown to be two important tasks~\cite{tang2021conceptguide} to improve the existing MOOC setting.


\subsection{Concept Pre-requisite Prediction }

Concept pre-requisite prediction is the task of identifying the pre-requisite relations between different concept. Once prerequisite relations among concepts are learned, these relations
can be used to prepare concept paths to guide  self-learners~\cite{tang2021conceptguide}. In addition, it can also help the course designers in designing course structure guided by the learned pre-requisites. Previous methods have exploited hand-crafted features and feed it to classification models~\cite{pan2017prerequisite} or investigate active learning with these features to improve classification models~\cite{liang2018investigating}. More recently, neural network based methods have been employed to classify pre-requisite relations such as,~\cite{roy2019inferring, li2019should}.  Among those~\cite{roy2019inferring} utilizes a Siamese network~\cite{bengio2013representation} which takes concept representation as input and predicts whether the first concept is pre-requisite of the other. On the other hand,~\cite{li2019should} considers the problem of pre-requisite prediction as link prediction task with concepts as nodes of graph and utilizes Graph Variational Autoencoder~\cite{kipf2016variational}. In our work, we learn pre-trained embeddings of the concepts using only their textual content and the structural relations derived from the MOOC platform.

\subsection{Lecture Recommendation}
Lecture recommendation aims at recommending relevant lectures to learners based on their  historical access behaviors. Such next lecture to watch recommendations
as a service has been shown to stimulate and excite learners when they are bored. The overwhelming selection of possible next steps in a MOOC compounded with the complexity of course content can leave a learner frustrated; while, friendly next-step recommendation can be the support they need to move forward and persist~\cite{pardos2017enabling}. To provide this service, researchers have looked into  of various methods. Pardos et. al.~\cite{pardos2017enabling} modeled the historic lectures that the student has watched as a sequence and uses sequence encoder to encode it and predict the next lecture of interest. The work in~\cite{gong2020attentional} takes as input the sequence of lectures that the student has watched along with the graph connecting various MOOC entities obtained from~\cite{yu2020mooccube}  to recommend the next lecture. This model employs an attention-based graph convolutional networks (GCNs) to learn the representation of different entities.  The model discovers learner potential interests by propagating learners’ preferences under the guide of meta-path in the graph. Some  studies extracted  hidden features from lectures (e.g., textual features extracted from lecture titles, visual features, and acoustic features) and used deep learning methods to learn their representation~\cite{xu2020course}. \par
Unlike these works, our model learns pre-trained embeddings of lectures using only the textual content and the course structure.
All these applications benefits the development of personalized
online learning system~\cite{yu2017towards}.
Our work provides a unified representation for MOOC entities compared with previous studies, and provides a solid
backbone for applications in the education domain.
\subsection{Pre-trained Representation Learning in NLP}
\begin{figure*}[!t]
   \centering
    {
       \includegraphics[width=0.90\textwidth]{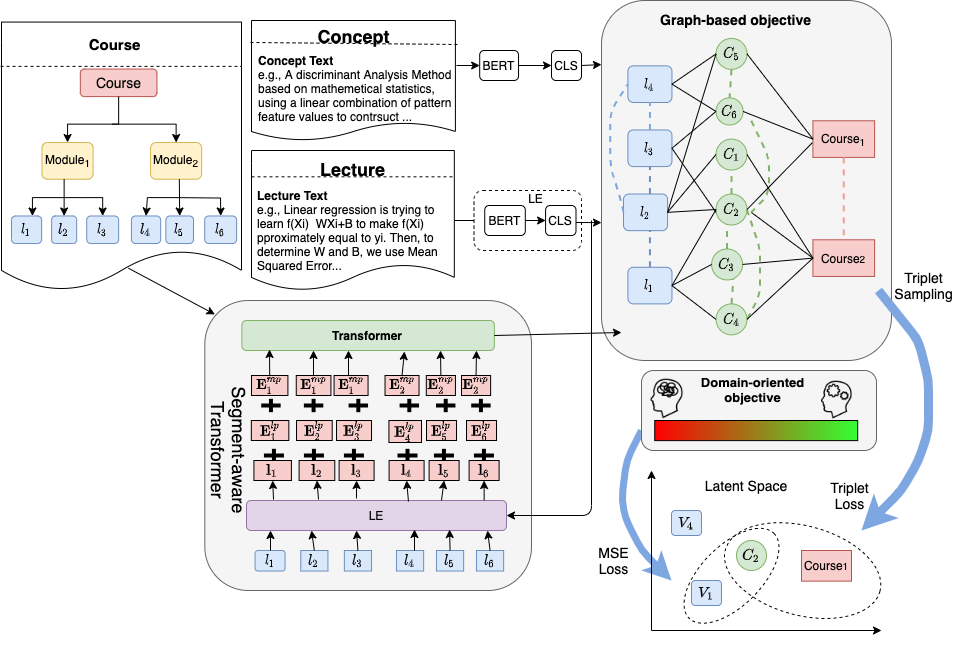}
     }
     \caption{The overview of MOOCRep model.}

     \label{architecture}
   \end{figure*}

Recent representation learning methods in NLP rely on training large neural language models on unsupervised data~\cite{peters2018deep, devlin2018bert, mccormick2016word2vec, lau2016empirical}.
together with some language related pre-training goals, solving
many NLP tasks with impressive performance. Although these pre-training solutions have been fully examined in a range of NLP tasks,
yet they are not effective to be directly applied to entity representation  mainly due to the following three reasons. First, MOOC entities in addition to having  textual features have structural relations between each other~\cite{yu2020mooccube}. These structural relations help in enhancing the embedding quality, would be ignored with these methods that only focus on text~\cite{cohan2020specter}. Second, deriving from domain knowledge, the design of courses by an instructor provide information about the entities other than just their linguistic features. Third, off-the-shelf application of these NLP approaches are difficult due to the need of model modification or hyper-parameter
tuning, which is inconvenient under many education setup.
In addition to that, methods have been developed to encode the relational knowledge in graphs~\cite{lerer2019pytorch}.  We use these models to generate embeddings of entities from the  graph shown in Figure ~\ref{architecture} and then use these embeddings as pre-trained embeddings for downstream tasks.
Our model learns pre-trained embeddings using multi-task objective designed to encode both the textual content and relational information between the entities.

\begin{table}[]
\caption{Notations}
\label{notations}
\begin{tabular}{ll}
\toprule
Notations & Description\\
\hline

$l_i$ & $i$th lecture in a course \\
$m_i$ & $i$th module in a course \\
$l$ & Maximum number of lectures in a course \\
$m$ & Maximum number of modules in a course \\
$\textbf{l}_i$ & Embedding of $i$th lecture in a course\\
$\textbf{z}_i$ & Embedding of $i$th course \\
$\textbf{c}_i$& Embedding of $i$th concept \\
$\textbf{E}^{lp}$ & Lecture position embedding matrix\\
$\textbf{E}^{mp}$ & Module position embedding matrix\\
$\hat{\boldsymbol{l}}_i$ & Position-aware embedding of $i$th lecture in a course\\
$\mathcal{G}$ &   Graph between MOOC entities   \\
$S_e$ & Set of negative sample edges for graph-based objective \\
$\mathcal{Z}_i$ & Set of courses associated with $i$th concept \\
$alc_i$ & Average lecture coverage of $i$th concept \\
$ast_i$ & Average survival time of $i$th concept \\
$d_i$ & Complexity level of $i$th concept\\

\bottomrule
\end{tabular}
\end{table}
\section{Proposed Work}
In this section, we introduce MOOCRep modeling and pre-training
in detail. First we give a formal definition of the MOOC entity representation problem. Then, we describe the MOOCRep architecture that learns MOOC entity embeddings. Afterwards we describe
the pre-training objectives we employed for training the MOOCRep model. Finally, in Section 3.4, we discuss how to apply the pre-trained embeddings to downstream tasks.
\subsection{Problem Formulation}
Here, we formally introduce the problem and describe the mathematical notations used in the paper. Mainly we use  unlabeled data from  MOOC structure. First structure originates from the hierarchical organization of MOOC courses. It usually consists of multiple modules, $\{m_1, m_2, \ldots, m_n \}$, and
each module consists of sequence of lectures, $\{ l_1, l_2, \ldots l_m\}$. For example, a course
on ``Computer Vision Basics'' has three modules on : ``Color, Light, and Image Formation'', ``Low-, Mid- and High-Level Vision
'', ``Mathematics for Computer Vision''. The
``Color, Light, and Image Formation'' has sequence of lectures: ``Light Sources'', ``Pinhole Camera Model'' and ``Color Theory''. Then, the second structure originates from the association between lectures and concepts.  Each lecture is cross linked with various concepts among $\{C_1, C_2, \ldots, C_l\}$ where each lecture covers multiple concepts and a concept can be covered by multiple lectures. For example, a lecture on ``Mathematics for Computer Vision'' is linked with the concept ``Basic Convex Optimization'' and ``Basic Convex Optimization'' can be associated with other lectures such as  ``Theory of Machine Learning''. In addition to these relation information, the textual content consisting of name and description of concepts and lectures is also available. \par
Our goal is to learn task-independent and effective representations of all the above described MOOC entities, i.e., courses, lectures, and concepts. Specifically, the output is a vector representation corresponding to each entity that encodes the individual entity feature along with different relations between them. In the following modules, we will address the main three challenges: (1) how to generate individual entity embeddings; (2) how
these embeddings are pre-trained; (3) how the embeddings are applied to downstream tasks. \par
The notations used in this paper have been described in table ~\ref{notations}
\subsection{Model Architecture}
In this subsection, we describe the architecture of MOOCRep model in detail. Figure~\ref{architecture} shows an overview of our model. The different components of MOOCRep are described below:
\subsubsection{Pre-trained BERT to encode the textual content}
To encode the textual contents of lectures and concepts including its name and description, we employ a pre-trained BERT-base model. Similar to the procedure employed for document representation \cite{beltagy2019scibert} the final representation of the [CLS] token is used as the output representation.
\begin{equation}
    \boldsymbol{x} = \text{BERT}(input)_{[CLS]},
\end{equation}
where BERT is the pretrained- BERT model \footnote{https://huggingface.co/bert-base-chinese}, and input is the concatenation of
the [CLS] token and WordPieces ~\cite{wu2016google}
of the name and description. We use the pre-trained BERT as our model initialization because it has been trained on large  Chinese simplified and traditional text.

\subsubsection{Encoding course structure with Segment-aware Transformers}
As shown in figure ~\ref{architecture}, a course  consists of an ordered sequence of modules which consists of lectures. This constitutes the hierarchical structure of courses and to learn effective course representation, it is important to take into account  this structural information. To this end, we employ \textit{segment-aware transformer} ~\cite{bai2020segabert} which has shown promising results at learning effective document representation from its hierarchical organization. It consists of an input embedding layer that generates position-aware  lecture embeddings and a transformer layer to aggregate the embeddings of the lecture sequence in the course to obtain the final course embedding. \\
\textit{Input Embedding Layer.}  To obtain an embedding of $i$th lecture in a course $c$, we first obtain the corresponding lecture representation using Equation (1). Then, we define a  \textit{lecture position} embedding matrix as $\textbf{E}^{lp} \in \mathbb{R}^{l \times d}$ to take into account the order of lectures present in the course, where $l$ is the maximum number of lectures in the course and \textit{module position} embedding matrix as $\textbf{E}^{mp} \in \mathbb{R}^{m \times d}$ to encode the position of module in the course, where $m$ is the maximum number of modules in the course.  Employing lecture position and module position encoding to learn lecture representation is important because it captures the order of lectures within a module and the order of modules in the course.  \par
To represent each lecture, we add the three embedding together: (1) the lecture embedding which captures the semantic features of each lecture,(2) the lecture's module position embedding , and (3) the lecture position embedding to obtain the position-aware lecture embedding as:
\begin{equation}
    \hat{\boldsymbol{l}}_{i} = \boldsymbol{l}_i +\boldsymbol{E}^{lp}_i + \boldsymbol{E}^{sp}_i,
\end{equation}
where $\boldsymbol{l}_i$ is obtained from Equation(1) using the textual content of the lecture.

Finally, the input lecture sequence is expressed as $\hat{\textbf{L}}  = [\hat{\textbf{l}}_1, \hat{\textbf{l}}_2, \ldots \hat{\textbf{l}}_l]$ by combining the lecture embedding, the lecture position embedding and the module position embedding. \par
\textit{Transformer Layer}
To learn a course  embedding, we take the input lecture sequence $\hat{\textbf{L}}$ and add a special token- [CLS] at the end to represent the whole course.  Then we use transformer layer ~\cite{vaswani2017attention} to encode the entire sequence. The transformer layer is composed of two sub-layers:
\begin{align}
    \boldsymbol{h}_l=\text{LayerNorm}(\boldsymbol{z}_{l-1}+\text{MHAtt}(\boldsymbol{z}_{l-1})),\\
    \boldsymbol{z}_l = \text{LayerNorm}(\boldsymbol{h}_l+\text{FFN}(\boldsymbol{h}_{l})),
\end{align}
\noindent where LayerNorm is a layer normalization proposed in ~\cite{ba2016layer} ; MHAtt is the multihead attentionmechanism introduced in ~\cite{vaswani2017attention} which allows each element of the sequence to attend to other elements with different attention weights; and FFN is feed-forward network with ReLU as the activation function. We take the course [CLS] token representation output
by the last layer of the transformer to represent the
course embedding, denoted as $\boldsymbol{z}_c$. Since the lecture embeddings used as input to the transformer layer are position-aware, i.e., they have already incorporated the hierarchical organization, the learned course embeddings will also be position-aware.   \par

The overall parameters of MOOCRep model includes 1) BERT model parameters, and 2) parameters of the segment-aware transformer. To learn these parameters we design the loss functions described in the next section to train the model in an end-to-end fashion.
\subsection{Pre-training Objectives}
In this subsection, we derive the objectives on which we learn our MOOCRep model. Specifically, these objectives are designed to capture the graph structure existing between the MOOC entities (graph-based objective) and the domain-oriented knowledge about the complexity of a concept (domain-oriented objective).
\subsubsection{Graph based  objective }
The graph between lectures, concepts and courses, shown in Figure ~\ref{architecture}, can help in identifying the related entities. To encode this relatedness signal, we design objective functions to train the MOOCRep model so that related entities lie closer in the embedding space. Particularly, our graph from ~\cite{yu2020mooccube} consists of vertices of types concepts, lectures, and courses. The edges connecting these vertices can be of two types:
\begin{itemize}
    \item \textbf{Explicit relation} In the lecture-concept and course-concept bipartite graphs, edges exist between lectures and concepts and course and concepts, presenting an explicit signal as it is directly available in the dataset.
    \item \textbf{Implicit relation} This relation indicates the similarity between the entities of the same type and can be induced from the  graph but are not provided explicitly. Specifically, if the number of adjacent nodes between two vertices increases by a threshold, then we consider an implicit relation between those vertices.
\end{itemize}

For learning representation of each node in the graph, MOOCRep optimizes a margin-based ranking objective between each edge $e$ in the training data and a set of edges $e'$ constructed by corrupting $e$.
\begin{equation}
    \mathcal{L}_{triplet} = \sum_{e \in \mathcal{G}} \sum_{e' \in S_e} \max(f(e)-f(e') + \lambda ,0)
\end{equation}
where $\lambda$ is a margin hyperparameter; $f$ is the cosine similarity between the embeddings of two nodes connected by that edge, i.e., $
    f(e=(s,d)) = cos(\boldsymbol{s},\boldsymbol{d})
$, where $\boldsymbol{s}$ and $\boldsymbol{d}$ correspond to the embedding of node $s$ and node $d$, respectively; $S_e$ is the set of negative samples obtained by corrupting each edge $e \in \mathcal{G}$  \par
\begin{equation}
    S_e = \{(s, d')\forall \{e= (s,d) \in \mathcal{G}\ \cap (s,d') \notin \mathcal{G} \cap d'\in type(d)\} \},
\end{equation}
\noindent where $type(d)$ returns the type of vertex $d$, lecture, or course.
\subsubsection{Domain-Oriented Objective}
Through pre-trained BERT model and graph-based objective, we have taken into account the relatedness between entities based on their textual content as well as connectivity in the graph, respectively. However, in addition to relatedness an important factor to consider is the ordering of entities based on their complexity levels. For example, two concepts, \textit{trees} and \textit{binary search trees} are relevant, but one (\textit{trees}) is a basic concept compared to (\textit{binary search trees}).  In order to include such information in final embeddings, we design a domain-oriented objective for our MOOCRep model. We predict the complexity level of entities from their embeddings and match it with their actual complexity level. \par

In the absence of ground-truth complexity levels of entities, we utilize heuristic methods to compute them. The distributional pattern of concepts over MOOCs convey information about the complexity of concepts.  The idea is that when instructors design  courses they explain the basic concepts first and teach an advanced concept by using the basic concepts they have already taught. Intuitively, for a concept if it covers more lectures in a course (lecture coverage) or survives longer time in a course (survival time), then it is more likely to be a basic concept rather than
an advanced one ~\cite{pan2017prerequisite}. We then use the following formal equations to compute  average lecture coverage ($alc$) and the average
survival time ($ast$) of a concept $i$ as follows,
\begin{equation}
    alc_i= \frac{1}{|\mathcal{Z}_i|} \sum_{z \in \mathcal{Z}_i} \frac{|I(z,i)|}{|z|},
    \end{equation}
    \begin{equation}
    ast_i = \frac{1}{|\mathcal{Z}_i|}\sum_{z \in \mathcal{Z}_i} \frac{|max(I(z,i)) - min(I(z,i))+1|}{|z|},
\end{equation}
\noindent where $\mathcal{Z}_i$ is the set of courses associated with $i$th concept, $z$ represents a course as an ordered sequence of lectures present in it.  The final complexity level of concept $i$ is determined as mean of $alc_i$ and $ast_i$, $d_i= \frac{(alc_i+ast_i)}{2}$. \par
To preserve the complexity level information in the learned embeddings,  we use a linear layer to map the embeddings $\boldsymbol{e}_i$ to a difficulty approximation $\hat{d}_i = \boldsymbol{w}_d^T \boldsymbol{e}_i + \boldsymbol{b}_d$ where $\boldsymbol{w}_d$ and $\boldsymbol{b}_d$
are network parameters. We use the computed complexity level $\boldsymbol{d}_i$ as
the auxiliary target, and design the following loss function $\mathcal{L}_{mse}$
to measure the approximation error:
\begin{equation}
    \mathcal{L}_{mse} = \sum_{i=1} ^C (d_i - \hat{d}_i)^2.
\end{equation}

Having designed two loss function, one to capture the graph relationships between entities and the other to preserve the complexity of the underlying concepts, we combine both the loss functions and obtain a multi-task objective as:
\begin{equation}
    \mathcal{L} = \lambda_1 \mathcal{L}_{triplet} + (1-\lambda_1) \mathcal{L}_{mse},
\end{equation}
\noindent where $\lambda_1$ is a  hyperparameter  to control the trade-off between graph-based objective and domain-oriented objective.

\subsection{MOOC Entity Embedding Evaluation}
After pre-training, MOOCRep entity embeddings can be directly applied as off-the-shelf features  to downstream tasks.  The first task we explore is concept pre-requisite prediction task, which can lead to  development of concept path that learners can use in deciding their learning path~\cite{tang2021conceptguide}. Such concept paths have nodes as concepts and directed edges indicating the pre-requisite relations between them.  We apply our learned embeddings from MOOCRep on state-of-the-art neural network based model ~\cite{roy2019inferring}. It essentially, learns concept embeddings using a Siamese network~\cite{bromley1993signature}. Another task important for education community is lecture recommendation ~\cite{pardos2017enabling, bhatt2018seqsense}. By introducing lecture recommendation based on learners interest, MOOC providers want to provide them the flexibility to access to a broader range of content~\cite{bhatt2018seqsense}. Neural network based methods have been proposed to solve this task. We apply MOOCRep embeddings to the model introduced in ~\cite{pardos2017enabling}, where lecture embeddings are learned from a sequence  encoder which is fed learners' lecture sequence as input. \par
To apply MOOCRep embeddings to a specific task,
we just provide the required embeddings as the entity's features to the downstream model, which minimizes the
cost of model modification. By doing this, we provide a better initialization to the downstream model, leading to their faster convergence and better optimization. In summary, MOOCRep has the following advantages for MOOC entity representation learning. First, it provides a unified and universally applicable embedding for MOOC entities. Second, it is able to incorporate both textual content and structural relation between entities along with the domain knowledge about complexity levels. Third, it is easy to directly  apply them on various downstream tasks.

\section{Experimental Settings}
In this section, we present our experimental settings to answer the following questions: \\
\textbf{RQ1} Can MOOCRep outperform the state-of-the-art methods for concept pre-requisite prediction and lecture recommendation tasks?  \\
\textbf{RQ2}: What is the influence of various components in the MOOCRep architecture?\\
\textbf{RQ3}: What does the qualitative analysis of MOOCRep embeddings reveal about their encoding behavior?\\
\textbf{RQ4}: What is the effect of training size of downstream tasks on the performance of  MOOCRep and other baselines?
\begin{table}[]
\centering
\caption{Dataset Details}
\begin{tabular}{l|r|r}

\toprule
                                    & \multicolumn{1}{c|}{Mathematics}                         & \multicolumn{1}{c}{Computer Science}                \\
                                    \midrule
\#Concepts                          & 373                          & 496                          \\
\#Courses                           & 459                          & 190                           \\
\#Lectures                            & 10,308                       & 5,085                       \\
\# Lectures/Course                    & 22.46                                             & 26.76                        \\
\#Tokens/Concept                    & 34.17                                             & 54.24                                          \\
\#Tokens/Lecture & 1365.73                                           & 1378.94                                 \\
\#Concepts/Course                   & 17.85 & 22.84 \\
\#Concepts/lecture                    & 1.93                                              & 18.19                                              \\
\# Concept Pre-requisites           & 1,314                                             & 1,604
\\
\#Users                             & 37,849                                            & 26,588                                         \\
\#Lecture/User                        & 60                                                & 67.79
\\
 \shortstack{\%Sequential elements\\ in interactions} & 62.1& 66.7\\

                                \bottomrule
\end{tabular}
\label{dataset}
\end{table}
\vspace{-2mm}
\subsection{Dataset} The dataset we used for our experiments have been obtained from the online education system called XuetangX and publicly available in~\cite{yu2020mooccube}. This dataset consists of concepts along with their description from Wikidata, lecture corpus from MOOCs, courses where each course consists of several modules and each module is covered by several lectures.  The lectures are  annotated with set of concepts . We consider the data from two domains, Mathematics and Computer Science.
The details for the dataset are shown in Table ~\ref{dataset}. \par

There are total $373$ and $496$ concepts, $459$ and $190$ courses , and $10, 308$ and $5,085$ lectures  in Mathematics and Computer Science, respectively.
On average, in Mathematics domain each course has $22.46$ lectures and  $17.85$ concepts and each lecture has $1365.73$ tokens; while each concept has $34.17$ tokens. In Computer Science domain each course $26.76$ lectures and $22.84$ concepts; while each lecture has $1378.94$ token and each concept has $54.24$ tokens.  \par
Since the MOOC concepts are universal and our goals is to learn pre-trained embeddings of these concepts, we augment the concepts from other available datasets~\cite{pan2017prerequisite, li2019should, liang2018investigating} to our dataset.  This results in total $1,314$ pre-requisite relations for Mathematics concepts, while $1,604$ for Computer Science concepts. For lecture recommendation, we considered the user interaction with courses of Mathematics and Computer Science domain only. This results in $37,849$ and $26,588$ users, and $60$ and $67.79$ interactions per user on average for Mathematics domain and Computer Science domain, respectively. In order to verify that learners do not necessarily follow the sequence of lectures set by instructors, we also compute the percentage of times consecutive interaction patterns occur in the set sequence by instructors. We find that it only occurs $62.1\%$ and $66.7\%$ times in Mathematics and Computer Science dataset, respectively.
\vspace{-2mm}
\subsection{Evaluation Tasks}

 We employ state-of-the-art supervised learning methods for concept pre-requisite prediction~\cite{roy2019inferring} and lecture recommendation~\cite{pardos2017enabling}. To evaluate the  effectiveness of MOOCRep, we compare the quality of our pre-trained embeddings with  pre-trained embeddings learned from the competing approaches. All these methods are able to generate entity representation, and then be applied to the two models mentioned above as warm-initialization.  Specifically, these methods are: \par

\begin{itemize}
\item \textbf{Random:} We randomly assign the embeddings which essentially results in the original supervised models.
    \item \textbf{Word2vec:} Assign text tokens with the corresponding word2vec embedding~\cite{mikolov2017advances} and CLS token as input to LSTM layer and use the embedding of [CLS] token as the entity embedding.
\item \textbf{Doc2vec} Similar to word2vec but with an additional paragraph vector representing the~\cite{lau2016empirical}
\item \textbf{BERT} is a state-of-the-art pre-training method featuring bi-directional transformer layers and masked language model~\cite{devlin2018bert}.
\item \textbf{PBG} is an embedding system that takes only the graph as input  and learns entity representation in an unsupervised manner~\cite{lerer2019pytorch}. We specifically, employ TransE~\cite{wang2014knowledge} model to learn the representations.
\end{itemize}
\subsection{Implementation Details}
For each evaluation tasks, we split the dataset into $80\%$, $10\%$, and $10\%$ as training, validation, and test set.  We take pretrained BERT-base with
$12$ layers to encode local semantic features from lectures and concepts. Pretrained model weights are obtained from Pytorch transformer repository\footnote{https://huggingface.co/bert-base-chinese}.  Besides, we set the number of global
transformer layers as $2$ based on the preliminary experiments. We find the $2$ layer global transformer layers work much better than 1 layer. All transformer-based models/layers have $768$ hidden units. We set the epochs as $200$, batch size as $64$. For word2vec we used pre-trained word vectors from~\cite{qiu2018revisiting} and words are extracted using jieba ~\footnote{\url{https://github.com/jsrpy/Chinese-NLP-Jieba}}.  For doc2vec we used the gensim APIs to obtain document embedding~\footnote{https://www.tutorialspoint.com/gensim/gensim\textunderscore doc2vec\textunderscore .htm} . We kept the maximum number of tokens in the corpus as $512$ and $1024$ for concepts and lectures, respectively. If for any entity the length of corpus is larger, we truncate them; while if it is smaller we concatenate them with padding elements. Since pre-trained BERT model only takes documents with $512$ tokens,  for lectures, we run the BERT model twice, once using the first 512 tokens, then using the generated [CLS] embedding to initialize the first element in the next $512$ tokens.
For modeling transE, we used Pytorch BigGraph~\cite{lerer2019pytorch} to obtain entity embeddings with embedding size of $768$.

Th hyperparameters used in MOOCRep are described here. For segment-aware transformer model, we used hidden embedding size as $128$  and maximum length of lecture sequence $l$ in a course is $510$, while maximum length of module  sequence  $m$ is $65$. We run the model for $100$ epochs and use the entity  embeddings obtained in the last run as the final embeddings. The optimizer is Adam~\cite{kingma2014adam} with learning rate of $1e-3$.\par
We use an embedding size of $128$ for both the downstream tasks.
For concept pre-requisite prediction task, we used the code publicly available of PREREQ~\cite{roy2019inferring} on github.  We run the code for $100$ epochs to get the best results and use Adam optimizer with learning rate of $1e-5$.  For lecture recommendation task, we implemented RNN as we could not find any public available code. The code is run with $200$ epochs and use Adam optimizer with learning rate of $1e-3$. \par
While running the experiments, we fix the seed of random initialization preventing variations in results on different runs. All the codes are run on Nvidia GeForce GTX $1050$ Ti  GPU. Here is MOOCRep code: \url{https://github.com/codeanonymous/moocrep}.
\begin{table*}[]
\caption{Performance comparison. The best performing method  is boldfaced, and the second best method
in each row is underlined. Gains are shown in the last row.}
\label{comparison}
\begin{tabular}{l|rrr|rrr|rrr|rrr|r}
\toprule
         & \multicolumn{6}{c|}{Concept Pre-requisite Prediction}                                                                                                                                                                                                     & \multicolumn{6}{c|}{Lecture Recommendation}                                                                                                                                                                                          & \multicolumn{1}{l}{}                         \\
         \midrule
         & \multicolumn{3}{c|}{Mathematics}                                                                                  & \multicolumn{3}{c|}{Computer Science}                                                                             & \multicolumn{3}{c|}{Mathematics}                                                            & \multicolumn{3}{c|}{Computer Science}                                                       & \multicolumn{1}{l}{}                         \\
         \midrule
         & \multicolumn{1}{c}{P} & \multicolumn{1}{c}{R} & \multicolumn{1}{c|}{F1} & \multicolumn{1}{c}{P} & \multicolumn{1}{c}{R} & \multicolumn{1}{c|}{F1} & \multicolumn{1}{c}{HR@10}           & \multicolumn{1}{c}{nDCG@10}         & \multicolumn{1}{c|}{MRR}             & \multicolumn{1}{c}{HR@10}           & \multicolumn{1}{c}{nDCG@10}         & \multicolumn{1}{c|}{MRR}             & \multicolumn{1}{c}{Avg} \\
         \midrule
Random   & 0.583                 & 0.594                 & 0.587                  & 0.528                 & 0.571                 & 0.545                  & 0.417          & 0.297          & 0.278          & 0.372          & 0.262          & 0.247          & 0.440                                        \\
Word2vec & 0.630                 & 0.638                 & 0.634                  & 0.570                 & 0.577                 & 0.573                  & 0.458          & 0.358          & 0.344          & 0.438          & 0.325          & 0.298          & 0.487                                        \\
Doc2vec  & 0.645                 & 0.644                 & 0.642                  & 0.594                 & 0.606                 & 0.599                  & 0.598          & 0.476          & 0.451          & 0.556          & 0.427          & 0.400          & 0.553                                        \\
BERT     & \underline{0.646}           & \underline{0.660}           & \underline{0.652}            & \underline{0.630}           & \underline{0.616}           & \underline{0.621}            & 0.650          & \underline{0.598}    & 0.458          & 0.628          & 0.534          & 0.485          & 0.598                                        \\
PBG      & 0.636                 & 0.626                 & 0.630                  & 0.607                                      & 0.611                                      & 0.609                                       & \underline{0.654}                         & 0.592                               & \underline{0.552}                         & \underline{0.643}    & \underline{0.559}                         & \underline{0.502}                         & 0.602                                        \\
MOOCRep    & \textbf{0.701}        & \textbf{0.685}        & \textbf{0.692}         & \textbf{0.659}        & \textbf{0.670}        & \textbf{0.663}         & \textbf{0.683} & \textbf{0.615} & \textbf{0.604} & \textbf{0.679} & \textbf{0.614} & \textbf{0.605} & 0.656                                        \\
\midrule
Gain \%  & 8.557                                      & 5.771                                      & 6.162                                       & 8.567                                      & 9.656                                      & 8.867                                       & 4.404                               & 2.809                               & 9.402                               & 5.552                               & 9.857                               & 20.438                              & 8.970   \\
\bottomrule
\end{tabular}
\end{table*}

\section{Results and Discussion}
\subsection{Performance Comparison (RQ1)}
Our evaluation of MOOCRep’s pretrained entity representations on the two downstream tasks is shown in Table ~\ref{comparison}. Overall, we observe substantial improvements across both the tasks with average performance of $0.656$ across all metrics on all tasks which is a $8.97\%$  relative improvement over the next-best baseline. We now discuss the results in detail. \par
For concept pre-requisite prediction, we used PREREQ~\cite{roy2019inferring} as the base model and initialized the embeddings with  pre-trained embeddings from baseline methods and our MOOCRep model. Since concept pre-requisite prediction is a binary classification task, where given a pair of concepts $(a,b)$, the task is to predict whether $a$ is pre-requisite of $b$, we report Precision (P), Recall (R) and Macro-averaged F1 score (F1) to compare the different models.  We observe that PREREQ performance when trained on our representations is better than when trained on any
other baseline. Particularly, on the Mathematics (Computer Science) dataset, we obtain an $0.692 (0.663)$ F1 score which is about a relatively $6.16\% (8.87\%)$ relative improvement over the best baseline on each dataset respectively. \par

For lecture recommendation, we used the method proposed in~\cite{pardos2017enabling} as base model which employs an RNN to predict user's next lecture. We initialize the lecture embeddings with  those pre-trained from the baseline models and MOOCRep model. We use ranking metrics to evaluate the recommendation performance, specifically report HR@10, nDCG@10 and MRR. We observe that MOOCRep outperforms all the baseline models on this task as well. For Computer Science dataset (Mathematics), MOOCRep achieves $0.683 (0.679), 0.615 (0.614),$ and $0.604 (0.605)$ at HR@10, nDCG@10, and MRR, respectively; MOOCRep outperforms by $4.40\% (5.52\%), 2.81\% (9.86\%),$ and $9.4\% (20.43\%)$ relative improvement over second best baseline on each dataset respectively. \par

Another observation is that BERT model which captures only the semantic features of  individual entities is the second best model for concept pre-requisite prediction; while PBG which captures the structural features between entities is the second best model for lecture recommendation. This can be attributed to the fact that the textual content of concepts is obtained from Wikipedia~\cite{yu2020mooccube} and hence the learned embeddings are good quality. On the other hand,  lectures' corpus is noisy resulting in BERT not able to generate their good quality embeddings. Our model, utilizing both textual content and structural relations between entities by taking advantage of 
\begin{table*}[]

\caption{Ablation Study}
\label{ablation}
\begin{tabular}{C{3.5cm}|rrr|rrr|rrr|rrr}
\toprule
                            & \multicolumn{6}{c|}{Concept Pre-requisite Prediction}                                                                                                                        & \multicolumn{6}{c}{Lecture Recommendation}                                                                                                                                      \\
                            \midrule
                            & \multicolumn{3}{c|}{Mathematics}                                 & \multicolumn{3}{c|}{Computer Science}                                                 & \multicolumn{3}{c|}{Mathematics}                                  & \multicolumn{3}{c}{Computer Science}                                                     \\
                            \midrule
                            & \multicolumn{1}{c}{P}      & \multicolumn{1}{c}{R}      & \multicolumn{1}{c|}{F1}     & \multicolumn{1}{c}{P}      & \multicolumn{1}{c}{R}      & \multicolumn{1}{c|}{F1}     & \multicolumn{1}{c}{HR@10}  & \multicolumn{1}{c}{nDCG@10} & \multicolumn{1}{c|}{MRR}    & \multicolumn{1}{c}{HR@10}  & \multicolumn{1}{c}{nDCG@10} & \multicolumn{1}{c}{MRR}    \\
                            \midrule

Without explicit similarity & 0.664 & 0.666 & 0.665 & 0.647 & 0.634 & 0.640 & 0.677 & 0.608  & 0.597 & 0.660 & 0.602  & 0.594                    \\
Without implicit similarity & 0.656 & 0.682 & 0.665 & 0.636 & 0.626 & 0.630 & 0.664 & 0.599  & 0.589 & 0.651 & 0.594  & 0.586                    \\
Without graph-based objective     & 0.649                      & 0.664                      & 0.653                      & 0.623 & 0.647 & 0.638 & 0.656 & 0.589  & 0.579 & 0.642                      & 0.593                       & 0.580                                   \\
Without domain-oriented objective  & 0.657 & 0.671 & 0.662 & 0.649 & 0.632 & 0.639 & 0.670 & 0.607  & 0.598 & 0.665 & 0.608  & 0.601                   \\
\hline
MOOCRep                       & \textbf{0.701} & \textbf{0.685} & \textbf{0.692} & \textbf{0.659} & \textbf{0.670} & \textbf{0.663} & \textbf{0.683} & \textbf{0.615}  & \textbf{0.604} & \textbf{0.679} & \textbf{0.614}  & \textbf{0.605}  \\
\bottomrule
\end{tabular}
\end{table*}
\subsection{Ablation Study (RQ2)}
\vspace{-2mm}
To get deep insights on the MOOCRep model, we investigate the contribution of various components involved in the model.
 Therefore, we conduct some ablation experiments to
show how each part of our method affects the final results. In Table ~\ref{ablation} ,
we compare the following variations of MOOCRep:
\begin{itemize}
    \item \textbf{Without explicit similarity} In this variant, we remove the explicit similarity objective when training the network. Specifically, we remove the course-concept and lecture-concept similarity loss.
    \item \textbf{Without implicit similarity} In this variant, we remove the loss function resulting from the implicit similarity between entities. Specifically, we remove the concept-concept, lecture-lecture, and course-course similarity loss.
    \item \textbf{Without graph-based objective} In this variant, we remove the triplet loss from the graph-based objective. The model is only trained with BERT followe by domain-oriented objective.
    \item \textbf{Without domain-oriented objective} In this variant, we remove the MSE loss from the domain-oriented objective (Equation 9). The model is traiend with BERT followed by graph-based objective.

\end{itemize}
The result in the above tables indeed shows many interesting conclusions.   Removing individual component does reduce the performance of methods on both concept pre-requisite prediction and lecture recommendation tasks.
\par
First the removal of loss resulting graph-based objective causes the most decline in the performance. Thus, incorporating the graph relationship between entities is the most important factor for MOOCRep. Second, the models show a similar degree of decline when removing explicit and implicit similarities of graph-based objective, which means these two pieces of
information are equally important.  The domain knowledge also plays a significant contribution to the model performance for concept pre-requisite prediction task; while not that much for the lecture recommendation task. The importance of concept complexity level for concept pre-requisite prediction task is that, for one concept to be pre-requisite of the other, it is important that they are both related and one is more difficult than the other. For lecture recommendation, the MOOCRep model takes into account the position of lecture in the course which already encodes the order relations between lectures in the same course. This information along with learner interaction information provided as training examples gives sufficient data for the lecture recommendation model to learn the ordering relations between the lectures. Thus explicit concept complexity does not inform the lecture recommendation task significantly.  \par
\begin{table*}[]

\centering
\caption{ Error cases for concept pre-requisite prediction on the test set. The textual content are not fully written for conciseness.}
\label{cases}
\resizebox{\textwidth}{!}{%
 \begin{tabular}{|C{6.5cm}|C{6.5cm}| C{1.2cm}C{0.5cm} C{0.5cm} C{1.2cm}|}

\toprule
Concept1 & Concept2 & Ground Truth &BERT & PBG & MOOCRep\\
\midrule
\textbf{Binary Tree:} A tree data structure. Each node has at most two subtrees, one is called the left subtree and the other is called the right subtree. & \textbf{Inorder Traversal:} A way to traverse a binary tree. The process can be expressed recursively as traversing the left subtree, then visiting the root node, and finally traversing the right subtree.& \ding[3]{51}& \ding[3]{51}&\ding[3]{51}&\ding[3]{51} \\
\hline
\textbf{Huffman Tree:} Among the binary trees with n weights as leaf nodes, the binary tree with the smallest weighted path length.  & \textbf{Minimum Spanning Tree:} For a weighted connected graph, the spanning tree with the smallest sum of weights on the edges.  &\ding[3]{55}  &\ding[3]{51} &\ding[3]{55} &\ding[3]{55} \\

\hline
\textbf{Divide and Conquer:} A complex problem is divided into multiple identical or similar sub-problems one by one, until the final sub-problem can be simply solved directly. The solution of the problem is the combination of the solutions of the sub-problems.&\textbf{Binary search:} A method for fast search on an ordered sequence. The main point is to compare the key value to be searched with the element> element located in the middle of the sequence each time.  &\ding[3]{51} &\ding[3]{55   } & \ding[3]{51}& \ding[3]{51} \\
\hline
\textbf{Quick Sort:} Divide the data to be sorted into two independent parts, such that all data in one part is smaller than all data in the other part, and then this method quickly sorts the two parts of data separately. & \textbf{Merge Sort:} The ordered sequence stored on an array is merged and sorted according to the odd and even address elements, the first step is to move the $>$ odd address elements to the left half \ldots& \ding[3]{55} & \ding[3]{51}&\ding[3]{51} &\ding[3]{55}  \\

\bottomrule
\end{tabular}}
\end{table*}
\vspace{-2mm}
\subsection{Qualitative Analysis of MOOCRep (RQ3)}
\vspace{-2mm}
To gain more insight, we analyze some cases of concept pre-requisite prediction task as shown in Table~\ref{cases}. The task of the downstream model was to predict if concept1 is pre-requisite of concept2. We compared the results with BERT, PBG and MOOCRep to generate pre-trained embeddings. Firstly, binary tree is pre-requisite of inorder traversal predicted correctly by all the four methods, as they have similar textual content and are linked to the same lecture. \par
To show the importance of modeling the graph relationships between the entities, we compare the results of Huffman tree and minimum spanning tree. BERT wrongly predicts the former to be pre-requisite of the latter as they have similar textual content. While, through structural relations, PBG and MOOCRep can identify that the two concepts are not pre-requisites.
On the other hand, divide and conquer is pre-requisite of  binary search, however their textual content is very dissimilar. Due to this, BERT can not identify the relevance between these concepts, while PBG and MOOCRep can derive the relations between them because these concepts are associated with the consecutively occurring lectures.\par
Lastly to show the benefit of encoding the complexity level of concept, we compare two sorting algorithms taught together in a lecture, quick sort and merge sort but do not have pre-requisite relations. Both BERT and PBG wrongly predict the pre-requisite relation due to similar content and relationship in the graph. However, MOOCRep that encodes the complexity levels of the concepts, correctly identifies that the two concepts are at the same complexity level, thus, projecting the embeddings of these concepts very close in the latent space. The distance between the two embeddings is not large enough for one to be pre-requisite of the other.
\subsection{Robustness to the proportion of training data  (RQ4)}
\vspace{-2mm}

   \begin{figure*}[!t]
   \centering
    {
       \includegraphics[width=0.8\textwidth]{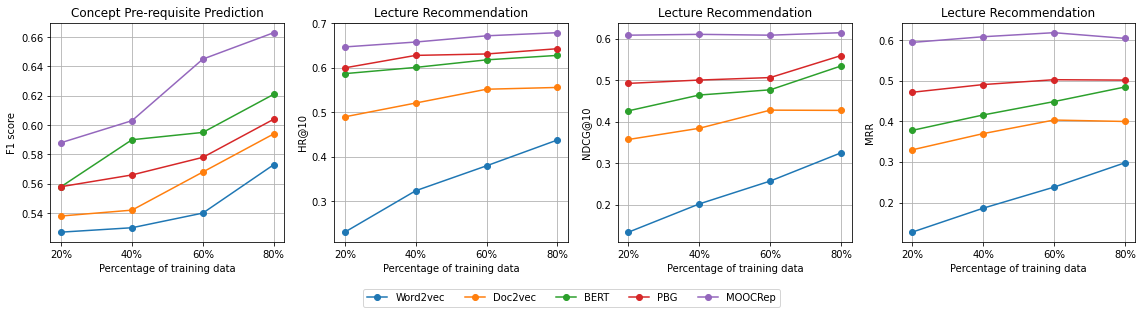}
     }

     \caption{Plot of performance of different evaluation tasks over different percentage of training data for each task.  Our model, MOOCRep significantly outperforms every baseline.}
      \label{performance}
   \end{figure*}
One benefit of exploiting the unlabeled data present in the MOOC structure is that it makes our model robust towards sparsity of data in the downstream tasks. To verify this, we vary the percentage of training data and compare the performance of the representation learning techniques on Computer Science dataset. \par
Specifically, we vary the training data percentage from $20\%$ to $80\%$. In each case, we take only $10\%$ of data samples  as validation and  $10\%$ as testing, respectively. This is done to compare the performance on the same testing data size. Figure~\ref{performance} shows how the performance of different methods change when varying the percentage of training data.
Our MOOCRep model significantly outperforms the competing approaches on both tasks when we vary the amount of training data. For the concept pre-requisite prediction task, we use the F1 score metrics. The performance of all the pre-training models is low when trained with $20\%$ and $40\%$ data. However, at only $60\%$ training data our model shows huge improvement compared to other models.   \par
 In addition, on the lecture recommendation task, MOOCRep shows better robustness, where the performance of other approaches degrades when percentage of training data is reduced.  Thus,we can reach to a conclusion that MOOCRep which exploits the relations between entities is effective in learning better entity embeddings even with constrained availability of training data in the downstream tasks.
\vspace{-2mm}
\section{Conclusion}
\vspace{-2mm}
In this paper, we addressed the problem of learning general-purpose pretrained embeddings of MOOC entities (lectures, concepts and courses). These embeddings not only encode the textual content of entities but also preserve MOOC structure and distributional pattern of concepts. To generate effective embeddings,  we leverage transformer model which is further pre-trained with the objectives designed to encode the graph relationships between entities and the complexity levels of concepts.  With extensive experiments on two important downstream tasks, concept pre-requisite prediction and lecture recommendation, we proved the superiority of MOOCRep embedings. We hope that the learned representation can be applied to other tasks important for the education community.\par
As part of future work, we plan to take into account other side information, such as instructor details and university offering the course. Secondly with the publicly available data about  ratings of different instructors and course given by learners, we can incorporate the information about quality of course and instructor in the embeddings. This information, will improve the quality of embeddings and enhance the performance on downstream applications, specifically the recommendation tasks.
\bibliographystyle{IEEEtran}

  \bibliography{ref.bib}

\begin{thebibliography}{10}
\providecommand{\url}[1]{#1}
\csname url@samestyle\endcsname
\providecommand{\newblock}{\relax}
\providecommand{\bibinfo}[2]{#2}
\providecommand{\BIBentrySTDinterwordspacing}{\spaceskip=0pt\relax}
\providecommand{\BIBentryALTinterwordstretchfactor}{4}
\providecommand{\BIBentryALTinterwordspacing}{\spaceskip=\fontdimen2\font plus
\BIBentryALTinterwordstretchfactor\fontdimen3\font minus
  \fontdimen4\font\relax}
\providecommand{\BIBforeignlanguage}[2]{{%
\expandafter\ifx\csname l@#1\endcsname\relax
\typeout{** WARNING: IEEEtran.bst: No hyphenation pattern has been}%
\typeout{** loaded for the language `#1'. Using the pattern for}%
\typeout{** the default language instead.}%
\else
\language=\csname l@#1\endcsname
\fi
#2}}
\providecommand{\BIBdecl}{\relax}
\BIBdecl

\bibitem{inoue2007online}
Y.~Inoue, \emph{Online education for lifelong learning}.\hskip 1em plus 0.5em
  minus 0.4em\relax IGI Global, 2007.

\bibitem{tang2021conceptguide}
C.-L. Tang, J.~Liao, H.-C. Wang, C.-Y. Sung, and W.-C. Lin, ``Conceptguide:
  Supporting online video learning with concept map-based recommendation of
  learning path,'' 2021.

\bibitem{bhatt2018seqsense}
C.~Bhatt, M.~Cooper, and J.~Zhao, ``Seqsense: video recommendation using topic
  sequence mining,'' in \emph{International Conference on Multimedia
  Modeling}.\hskip 1em plus 0.5em minus 0.4em\relax Springer, 2018, pp.
  252--263.

\bibitem{zhao2018flexible}
J.~Zhao, C.~Bhatt, M.~Cooper, and D.~A. Shamma, ``Flexible learning with
  semantic visual exploration and sequence-based recommendation of mooc
  videos,'' in \emph{Proceedings of the 2018 CHI Conference on Human Factors in
  Computing Systems}, 2018, pp. 1--13.

\bibitem{mahapatra2018videoken}
D.~Mahapatra, R.~Mariappan, V.~Rajan, K.~Yadav, and S.~Roy, ``Videoken:
  Automatic video summarization and course curation to support learning,'' in
  \emph{Companion Proceedings of the The Web Conference 2018}, 2018, pp.
  239--242.

\bibitem{roy2019inferring}
S.~Roy, M.~Madhyastha, S.~Lawrence, and V.~Rajan, ``Inferring concept
  prerequisite relations from online educational resources,'' in
  \emph{Proceedings of the AAAI Conference on Artificial Intelligence},
  vol.~33, no.~01, 2019, pp. 9589--9594.

\bibitem{liang2018investigating}
C.~Liang, J.~Ye, S.~Wang, B.~Pursel, and C.~L. Giles, ``Investigating active
  learning for concept prerequisite learning,'' in \emph{Proceedings of the
  AAAI Conference on Artificial Intelligence}, vol.~32, no.~1, 2018.

\bibitem{xu2016personalized}
J.~Xu, T.~Xing, and M.~Van Der~Schaar, ``Personalized course sequence
  recommendations,'' \emph{IEEE Transactions on Signal Processing}, vol.~64,
  no.~20, pp. 5340--5352, 2016.

\bibitem{mccormick2016word2vec}
C.~McCormick, ``Word2vec tutorial-the skip-gram model,''
  \emph{Apr-2016.[Online]. Available: http://mccormickml.
  com/2016/04/19/word2vec-tutorial-the-skip-gram-model}, 2016.

\bibitem{beltagy2019scibert}
I.~Beltagy, K.~Lo, and A.~Cohan, ``Scibert: A pretrained language model for
  scientific text,'' \emph{arXiv preprint arXiv:1903.10676}, 2019.

\bibitem{devlin2018bert}
J.~Devlin, M.-W. Chang, K.~Lee, and K.~Toutanova, ``Bert: Pre-training of deep
  bidirectional transformers for language understanding,'' \emph{arXiv preprint
  arXiv:1810.04805}, 2018.

\bibitem{bai2020segabert}
H.~Bai, P.~Shi, J.~Lin, L.~Tan, K.~Xiong, W.~Gao, and M.~Li, ``Segabert:
  Pre-training of segment-aware bert for language understanding,'' \emph{arXiv
  preprint arXiv:2004.14996}, 2020.

\bibitem{yu2020mooccube}
J.~Yu, G.~Luo, T.~Xiao, Q.~Zhong, Y.~Wang, J.~Luo, C.~Wang, L.~Hou, J.~Li,
  Z.~Liu \emph{et~al.}, ``Mooccube: A large-scale data repository for nlp
  applications in moocs,'' in \emph{Proceedings of the 58th Annual Meeting of
  the Association for Computational Linguistics}, 2020, pp. 3135--3142.

\bibitem{chechik2010large}
G.~Chechik, V.~Sharma, U.~Shalit, and S.~Bengio, ``Large scale online learning
  of image similarity through ranking,'' 2010.

\bibitem{lerer2019pytorch}
A.~Lerer, L.~Wu, J.~Shen, T.~Lacroix, L.~Wehrstedt, A.~Bose, and
  A.~Peysakhovich, ``Pytorch-biggraph: A large-scale graph embedding system,''
  \emph{arXiv preprint arXiv:1903.12287}, 2019.

\bibitem{anderson2014engaging}
A.~Anderson, D.~Huttenlocher, J.~Kleinberg, and J.~Leskovec, ``Engaging with
  massive online courses,'' in \emph{Proceedings of the 23rd international
  conference on World wide web}, 2014, pp. 687--698.

\bibitem{pan2017prerequisite}
L.~Pan, C.~Li, J.~Li, and J.~Tang, ``Prerequisite relation learning for
  concepts in moocs,'' in \emph{Proceedings of the 55th Annual Meeting of the
  Association for Computational Linguistics (Volume 1: Long Papers)}, 2017, pp.
  1447--1456.

\bibitem{pardos2017enabling}
Z.~A. Pardos, S.~Tang, D.~Davis, and C.~V. Le, ``Enabling real-time adaptivity
  in moocs with a personalized next-step recommendation framework,'' in
  \emph{Proceedings of the Fourth (2017) ACM Conference on Learning@ Scale},
  2017, pp. 23--32.

\bibitem{li2019should}
I.~Li, A.~R. Fabbri, R.~R. Tung, and D.~R. Radev, ``What should i learn first:
  Introducing lecturebank for nlp education and prerequisite chain learning,''
  in \emph{Proceedings of the AAAI Conference on Artificial Intelligence},
  vol.~33, no.~01, 2019, pp. 6674--6681.

\bibitem{bengio2013representation}
Y.~Bengio, A.~Courville, and P.~Vincent, ``Representation learning: A review
  and new perspectives,'' \emph{IEEE transactions on pattern analysis and
  machine intelligence}, vol.~35, no.~8, pp. 1798--1828, 2013.

\bibitem{kipf2016variational}
T.~N. Kipf and M.~Welling, ``Variational graph auto-encoders,'' \emph{arXiv
  preprint arXiv:1611.07308}, 2016.

\bibitem{gong2020attentional}
J.~Gong, S.~Wang, J.~Wang, W.~Feng, H.~Peng, J.~Tang, and P.~S. Yu,
  ``Attentional graph convolutional networks for knowledge concept
  recommendation in moocs in a heterogeneous view,'' in \emph{Proceedings of
  the 43rd International ACM SIGIR Conference on Research and Development in
  Information Retrieval}, 2020, pp. 79--88.

\bibitem{xu2020course}
W.~Xu and Y.~Zhou, ``Course video recommendation with multimodal information in
  online learning platforms: A deep learning framework,'' \emph{British Journal
  of Educational Technology}, vol.~51, no.~5, pp. 1734--1747, 2020.

\bibitem{yu2017towards}
H.~Yu, C.~Miao, C.~Leung, and T.~J. White, ``Towards ai-powered personalization
  in mooc learning,'' \emph{npj Science of Learning}, vol.~2, no.~1, pp. 1--5,
  2017.

\bibitem{peters2018deep}
M.~E. Peters, M.~Neumann, M.~Iyyer, M.~Gardner, C.~Clark, K.~Lee, and
  L.~Zettlemoyer, ``Deep contextualized word representations,'' 2018.

\bibitem{lau2016empirical}
J.~H. Lau and T.~Baldwin, ``An empirical evaluation of doc2vec with practical
  insights into document embedding generation,'' \emph{arXiv preprint
  arXiv:1607.05368}, 2016.

\bibitem{cohan2020specter}
A.~Cohan, S.~Feldman, I.~Beltagy, D.~Downey, and D.~S. Weld, ``Specter:
  Document-level representation learning using citation-informed
  transformers,'' in \emph{Proceedings of the 58th Annual Meeting of the
  Association for Computational Linguistics}, 2020, pp. 2270--2282.

\bibitem{wu2016google}
Y.~Wu, M.~Schuster, Z.~Chen, Q.~V. Le, M.~Norouzi, W.~Macherey, M.~Krikun,
  Y.~Cao, Q.~Gao, K.~Macherey \emph{et~al.}, ``Google's neural machine
  translation system: Bridging the gap between human and machine translation,''
  \emph{arXiv preprint arXiv:1609.08144}, 2016.

\bibitem{vaswani2017attention}
A.~Vaswani, N.~Shazeer, N.~Parmar, J.~Uszkoreit, L.~Jones, A.~N. Gomez,
  L.~Kaiser, and I.~Polosukhin, ``Attention is all you need,'' \emph{arXiv
  preprint arXiv:1706.03762}, 2017.

\bibitem{ba2016layer}
J.~L. Ba, J.~R. Kiros, and G.~E. Hinton, ``Layer normalization,'' \emph{arXiv
  preprint arXiv:1607.06450}, 2016.

\bibitem{bromley1993signature}
J.~Bromley, I.~Guyon, Y.~LeCun, E.~S{\"a}ckinger, and R.~Shah, ``Signature
  verification using a" siamese" time delay neural network,'' \emph{Advances in
  neural information processing systems}, vol.~6, pp. 737--744, 1993.

\bibitem{mikolov2017advances}
T.~Mikolov, E.~Grave, P.~Bojanowski, C.~Puhrsch, and A.~Joulin, ``Advances in
  pre-training distributed word representations,'' \emph{arXiv preprint
  arXiv:1712.09405}, 2017.

\bibitem{wang2014knowledge}
Z.~Wang, J.~Zhang, J.~Feng, and Z.~Chen, ``Knowledge graph embedding by
  translating on hyperplanes,'' in \emph{Proceedings of the AAAI Conference on
  Artificial Intelligence}, vol.~28, no.~1, 2014.

\bibitem{qiu2018revisiting}
Y.~Qiu, H.~Li, S.~Li, Y.~Jiang, R.~Hu, and L.~Yang, ``Revisiting correlations
  between intrinsic and extrinsic evaluations of word embeddings,'' in
  \emph{Chinese Computational Linguistics and Natural Language Processing Based
  on Naturally Annotated Big Data}.\hskip 1em plus 0.5em minus 0.4em\relax
  Springer, 2018, pp. 209--221.

\bibitem{kingma2014adam}
D.~P. Kingma and J.~Ba, ``Adam: A method for stochastic optimization,''
  \emph{arXiv preprint arXiv:1412.6980}, 2014.

\end{thebibliography}

\end{document}